\documentclass[]{article}
\usepackage[a4paper]{geometry}
\usepackage{mtsummit2023}
\usepackage{times}
\usepackage{url}
\usepackage{latexsym}
\usepackage{natbib}
\usepackage{layout}
\usepackage[hang,marginal]{footmisc}
\usepackage{amsmath}
\usepackage[capitalize,noabbrev]{cleveref}
\usepackage{graphicx}
\usepackage{mathtools}

\usepackage[utf8]{inputenc}
\usepackage[english]{babel}
\usepackage{xcolor}

\usepackage[normalem]{ulem} 

\DeclareMathOperator*{\margin}{margin}
\DeclareMathOperator*{\xsim}{xsim}
\DeclareMathOperator*{\NN}{NN}

\DeclareMathOperator*{\dec}{dec}
\DeclareMathOperator*{\enc}{enc}
\DeclareMathOperator*{\PseudoPar}{\text{\textit{PseudoPar}}} 
\begin{document}

\title{\bf Boosting Unsupervised Machine Translation\\ with Pseudo-Parallel Data}

\author{\name{\bf Ivana Kvapilíková} \hfill  \addr{kvapilikova@ufal.mff.cuni.cz}\\
        \name{\bf Ondřej Bojar} \hfill \addr{bojar@ufal.mff.cuni.cz}\\
        \addr{Charles University, Faculty of Mathematics and Physics, Institute of Formal and Applied Linguistics, Prague, 118 00, Czechia}
}


\maketitle
\pagestyle{empty}

\begin{abstract}
Even with the latest developments in deep learning and large-scale language modeling, the task of machine translation (MT) of low-resource languages remains a challenge.  Neural MT systems can be trained in an unsupervised way without any translation resources but the quality lags behind, especially in truly low-resource conditions. We propose a training strategy that relies on pseudo-parallel sentence pairs mined from monolingual corpora in addition to synthetic sentence pairs back-translated from monolingual corpora. We experiment with different training schedules and reach an improvement of up to 14.5 BLEU points (English to Ukrainian) over a baseline trained on back-translated data only.


\end{abstract}



\section{Introduction}


After the great advancements in machine translation (MT) quality brought by neural MT (NMT; \citealp{bahdanau,transformer}) trained on millions of pre-translated sentence pairs, there came a realization that parallel data is expensive and surely not available for most language pairs in the world. Researchers started focusing their attention on methods leveraging monolingual data for machine translation \citep{sennrich} and even explored the extreme scenario of training a translation system in a completely unsupervised way with no parallel data at all \citep{artetxe2018nmt,lample2018only}. 

The recent impressive progress in language modeling did not leave the area of machine translation intact. However, the translation capabilities of large language models such as the latest GPT models \citep{gpt} are weak for underrepresented languages \citep{hendy2023good} and unsupervised MT aimed at low-resource languages still deserves special attention.


There are two ways to approach machine translation trained exclusively on monolingual data. In the absence of parallel texts, the monolingual training sentences can either be coupled with their synthetic counterparts which are automatically generated through back-translation \citep{artetxe2018nmt,lample2018only}, or with authentic counterparts which are automatically selected from existing monolingual texts to be as close translations as possible \citep{ruiter-etal-2019-self}. Researchers have successfully explored both of these avenues with the conclusion that it is indeed possible to train a functional MT system on monolingual texts only. However, little attention has been paid to combining the two approaches together.

In this paper, we work with the standard framework for training unsupervised MT but we  incorporate an additional training step where sentence pairs mined from monolingual corpora are used to train the model with a standard supervised MT objective. We consider the mined sentence pairs as \textit{pseudo-parallel} as they should ideally be identical in meaning but in practice only share a certain degree of similarity. We show that they improve the translation quality nonetheless. We experiment with different training schedules to determine when to incorporate the pseudo-parallel data and when to remove it from the training.

In \cref{sec:relwork}, we summarize the related work on the topics of unsupervised MT and parallel corpus mining. In \cref{sec:meth}, we introduce our method, focusing on how we obtain the pseudo-parallel sentences and how we incorporate them into the unsupervised MT training. \cref{sec:exp} gives the results of our experiments which are discussed in \cref{sec:disc}.


\section{Related Work}
\label{sec:relwork}

\subsection{Unsupervised MT}
Unsupervised MT was first tackled by \citet{artetxe2018nmt} and \citet{lample2018only} who introduced a neural model with shared encoder parameters for both language directions that was capable of translating without being trained on parallel data. The authors relied on pre-trained embeddings to ignite the learning process and then trained the model using denoising \citep{dae} and back-translation \citep{sennrich-etal-2016-improving}. \citet{artetxe2018smt} and \citet{lample2018only} also explored the possibilities of unsupervised phrase-based MT where the initial phrase table is induced from a cross-lingual embedding space. 

A significant improvement in neural models was brought by splitting the training of the entire model into a pre-training phase where the weights are first trained on an auxiliary task aimed at language understanding (e.g. masked language modeling, denoising) and a fine-tuning phase where the model is trained for translation. \citet{conneau2019pretraining} train a cross-lingual BERT-like \citep{BERT} language model on the concatenation of the monolingual corpora and copy its weights to initialize the parameters of both the encoder and the decoder. 
\citet{song2019mass} reach slightly better translation quality by pre-training the entire sequence-to-sequence model to reconstruct a missing piece of a sentence given the surrounding tokens. 

\citet{liu-etal-2020-multilingual-denoising} explore the benefits of multilingual pre-training of the entire translation model on the task of multilingual denoising (mBART) and reach state-of-the-art results in unsupervised MT. \citet{ustun-etal-2021-multilingual} extend the pre-trained mBART model with denoising adapters and fine-tune on auxiliary parallel language pairs without the need for back-translation. \citet{garcia2020multilingual, garcia-etal-2021-harnessing} train a multilingual translation system and combine back-translation from monolingual data with cross-translation of auxiliary parallel data in high-resource language pairs.

 Unsupervised MT has been influenced by the latest advancements in large-scale multilingual language modeling \citep{nllb2022}. The GPT-3 model \citep{gpt} is capable of translation without being trained on an explicit translation objective and its performance increases considerably with one-shot or few-shot fine-tuning. However, its ability to handle low-resource and non-English-centric language pairs lags behind \citep{hendy2023good}.


\subsection{Parallel Corpus Mining for MT}

Using mined sentence pairs for MT training was heavily explored by \citet{schwenk-2018-filtering} and \citet{Artetxe2019laser} who introduced LASER, a multilingual sentence encoder that is able to find translation equivalents in 93 languages with high precision. \citet{nllb2022} extend the approach to cover 200 languages by student-teacher training. 
However, the training of the teacher model is heavily supervised by millions of parallel sentence pairs and its distillation also requires at least some parallel sentences.

\citet{ruiter-etal-2019-self} introduce self-supervised translation where the model used for selecting translation examples is the emergent NMT model itself. The authors search for the nearest neighbors in a sentence embedding space extracted from an NMT system and apply a strong filter to only select meaningful candidates for training. \citet{criss} use self-supervised training of a pre-trained multilingual model (mBART) which iteratively selects parallel sentence pairs and trains itself on the mined examples. They show an improvement over the mBART model fine-tuned on back-translated data only.

Similar to our work, \citet{ruiter-etal-2021-integrating} incorporate a training step using denoising and back-translation into their self-supervised MT system. We take the opposite direction to reach a similar goal when we start from an unsupervised MT system and incorporate a training step supervised by the mined sentence pairs extracted outside of the NMT model. \citet{kvapilikova-bojar-2022-cuni} observed a positive role of pseudo-parallel data in an unsupervised MT shared task but the most effective way to integrate this type of data into the training is yet to be established. 








\section{Unsupervised MT with Pseudo-Parallel Data}
\label{sec:meth}

It was demonstrated by \citet{artetxe2018nmt} and
 \citet{lample2018only} that the key elements of an unsupervised neural MT are shared model parameters, good initialization, and iterative learning on back-translated data. We build upon the existing work in unsupervised MT and extend the training procedure with a training step leveraging pseudo-parallel sentence pairs obtained from monolingual training corpora.

\subsection{Search for Pseudo-Parallel Data}

A multilingual language model trained on monolingual data only can be used to create language-neutral sentence representations \citep{libovicky-etal-2020-language} in an unsupervised way. Pseudo-parallel sentence pairs are retrieved as closest neighbors in the multilingual space \citep{schwenkartexte2018}.

\subsubsection*{Sentence Encoder}
Multilingual masked language models (MLMs) such as mBERT \citep{BERT}, XLM \citep{conneau2019pretraining} and XLM-R \citep{conneau2019unsupervised_at_scale} are Transformer \citep{transformer} encoders trained with a masked language modeling (MLM) objective \citep{BERT} where random tokens from the input text stream are masked and the model is trained to predict them back. MLM models create representations where each token carries information about its left and right context.  Sentence embeddings can be retrieved from any layer of the model but the per-token encoder outputs need to first be aggregated, e.g. by taking their mean or their element-wise maximum over the sentence tokens.

\citet{pires2019} and \citet{libovicky-etal-2020-language} studied the language neutrality of the representations produced by multilingual language models and \citet{kvapilikova2020} showed that with minimal fine-tuning, the sentence embeddings extracted from the mid-layers of the model by mean-pooling per-token encoder outputs can be used for parallel corpus mining. They also observed that fine-tuning an MLM sentence encoder on a small synthetic parallel corpus increases both precision and recall on the task of parallel sentence mining even for unrelated language pairs. 




\subsubsection*{Parallel Sentence Search}

To perform the search for parallel sentence pairs, all sentences from the two monolingual corpora are encoded and all possible sentence combinations are scored to select the most similar sentence pairs. The scoring is performed by a margin-based similarity metric \citep{schwenkartexte2018}

\begin{equation}
\xsim(x, y) = \margin\Bigl(\cos(x, y),\sum_{z\in \NN_k(x)}\frac{\cos(x, z)}{2k} + \sum_{z\in \NN_k(y)}\frac{\cos(y, z)}{2k}\Bigr)
\label{eq:xsim}
\end{equation}

where $\margin(a,b)=\frac{a}{b}$, $\NN_k(x)$ is the set of $k$ nearest neighbors of $x$. The method for scoring involves cosine similarity which is comparatively evaluated against the average cosine similarity of a given sentence with its nearest neighbors to eliminate the ``hubs''. When the score surpasses a designated threshold $T$, two sentences are deemed to be parallel:

\begin{equation}
\text{xsim}(x, y) > T
\label{eq:thresh}
\end{equation}

\subsection{Unsupervised MT Architecture}

The design of an NMT system needs to meet several requirements to be functional for unsupervised translation. Firstly, a significant number of parameters needs to be shared among the languages in order to allow the model to generate a shared latent space where meaning is represented regardless of the language it is expressed in \citep{lample2018}. Secondly, the initialization of the model weights is vital to produce an initial solution and kick-start the training process \citep{conneau2019pretraining}.

The configuration of our unsupervised MT system follows that of \citet{conneau2019pretraining} and consists of a Transformer 
encoder and decoder, both of which are shared  between the two languages. The tokenized input in both languages is processed by a single BPE \citep{sennrich} model learned on the concatenation of the two monolingual corpora and the joint vocabulary enables both languages to use a shared embedding matrix. 


\subsection{Unsupervised Pre-Training}

The model is initialized with weights from a masked language model pre-trained on the monolingual corpora and copied into both the encoder and the decoder as in \citet{conneau2019pretraining}. The initialized model is further pre-trained as a bilingual denoising autoencoder \citep{liu-etal-2020-multilingual-denoising}. The fine-tuning of the pre-trained model is scheduled in stages which are discussed in \cref{ssec:fine}. 

\subsection{Fine-Tuning for Translation}
\label{ssec:fine}
The pre-trained model is fine-tuned on both back-translated and pseudo-parallel data which are combined into different training schedules to determine their role at a given point in training. Intuitively, non-equivalent sentence pairs with some translation information should be useful at the beginning of the training when the model has minimal or no cross-lingual information. However, as the training progresses, it starts to produce synthetic translations of increasing quality which at a certain point surpass the quality of the pseudo-parallel corpus. We hypothesize that the most effective approach is to train the model on both synthetic and pseudo-parallel data until a certain breaking point, and from that point on, continue training solely on synthetic data.

\subsubsection{Fine-Tuning on Pseudo-Parallel Data}
To fine-tune the model on pseudo-parallel data, the standard supervised MT objective is used. In every step of the training, a mini-batch of pseudo-parallel sentences is added and the model is trained to minimize the loss function

\begin{equation}
    L_{PPMT}(\theta_{\enc},\theta_{\dec}) = \\
    E_{(x,y)\sim \PseudoPar,\hat{y}\sim \dec(\enc(x))}\Delta(\hat{y},y)
\end{equation}

where $(\theta_{\enc},\theta_{\dec})$ is the trained model, ($x,y$) is a sentence pair sampled from the pseudo-parallel data set $PseudoPar$, and $\Delta$ is the cross-entropy loss.

\subsubsection{Fine-Tuning on Iteratively Back-Translated Data}
In the back-translation step, the model is first set to the inference mode and used to translate a batch of sentences. The synthetic translations serve as source sentences fed into the model while the original sentences serve as the ground truth for the cross-entropy loss computation. The back-translation loss for translation from language $Lsrc$ to $Ltgt$ is defined as 
\begin{equation}
    L_{IBT}(\theta_{\enc},\theta_{\dec}, Ltgt) = \\
    E_{x\sim D_{Ltgt},\hat{x}\sim \dec(\enc(T(x))}(\Delta(\hat{x},x))
\end{equation}

where $x$ is a sentence sampled from the target corpus $D_{Ltgt}$, $T(x)$ is the translation model which generates a synthetic translation of $x$, and $\Delta$ is the cross-entropy loss.

\section{Experimental Details}
\label{sec:exp}
\subsection{Data}


       
    

\begin{table}[]
    \centering
      \small
    \begin{tabular}{l|c|c|c|c}
       &  \textbf{de-hsb} & \textbf{en-ka} & \textbf{en-kk} & \textbf{en-uk}  \\
       \hline
        train (mono)  &29.4M/0.9M  &17.1M/6.6M & 17.1M/7.7M & 17.1M/17.3M\\
       \hline
      train (pseudo-parallel) & 770K & 230K & 169K & 496K \\
         
    \end{tabular}
    \caption{Number of sentences in the monolingual corpora and mined pseudo-parallel corpora. }
    \label{tab:data}
\end{table}

We train translation models for the following language pairs: German-Upper Sorbian (de-hsb), English-Georgian (en-ka), English-Kazakh (en-kk) and English-Ukrainian (en-uk). The German and Upper Sorbian monolingual training data as well as the parallel validation and test sets were provided in the WMT22 unsupervised shared task \citep{wellerdimarco-fraser:2022:WMT2}. The monolingual training data for the other languages come from the Oscar\footnote{\url{https://oscar-project.org/}} corpus. The training data summary is given in \cref{tab:data}. The English-centric validation and test sets were taken from the Flores Evaluation Benchmark \citep{nllb2022}. In addition, the legal test sets from the MT4All shared task \citep{de-gibert-bonet-etal-2022-unsupervised_mt4all} were used for evaluation. 

The data was tokenized and split into BPE units using the fastText \citep{joulin2016fasttext} library. We shared one BPE vocabulary of 55k entries for en-ka-kk-uk and another vocabulary of 18k entries for de-hsb. 

\subsection{Training Details}
\subsubsection{Model Architecture} 
All our translation models have a dual character to translate in both translation directions. They have the same 6-layer Transformer 
architecture with 8 attention heads and the hidden size of 1024, language embeddings, GELU \citep{gelu} activations and a dropout rate of 0.1. For language model pre-training, we use mini-batches of 64 text streams (256 tokens per stream) per GPU and Adam \citep{adam} optimization with \texttt{lr=0.0001}. For denoising and MT fine-tuning, we use mini-batches of 3400 tokens per GPU and Adam optimization with a linear warm-up \texttt{(beta1=0.9,beta2=0.98,lr=0.0001)}. The models are trained on 8 GPUs. We use the XLM\footnote{\url{https://github.com/facebookresearch/XLM}} toolkit for training.


 \noindent
\subsubsection{Sentence Encoder}

We use the  XLM-100 model \citep{conneau2019pretraining} fine-tuned on English-German synthetic sentence pairs according to \citet{kvapilikova2020} as our sentence encoder. To measure its ability to create representations with a high level of multilingualism, we evaluate its performance of an auxiliary task of parallel sentence mining (PSM). For each language pair, we randomly select 200k sentences from the monolingual data, mix in the parallel validation set, and measure the precision and recall of the model when trying to reconstruct it.

Since XLM-100 was trained on 100 languages and Upper Sorbian is not one of them, we fine-tune the model on German and Upper Sorbian sentences before using it to mine parallel sentence pairs. We stop fine-tuning when the quality of the mined corpus starts deteriorating. We determine the optimal length of fine-tuning on the PSM task and observe that both precision and recall start slowly decreasing after the model had seen 500k sentences.

To retrieve sentence embeddings from the trained model, we mean-pool the encoder outputs from the fifth-to-last layer across sentence tokens (the layer and aggregation choice follow \citet{kvapilikova2020}). We search the embedding space as described in \cref{eq:xsim} and \cref{eq:thresh}. We select a threshold $T$ that maximizes the F1 score on the PSM task. \cref{tab:psm} lists the precision and recall of all sentence encoders used for mining together with the optimal mining threshold. The amount of mined parallel sentences used for unsupervised MT training is given in \cref{tab:data}.



\begin{table}[]
    \centering
      \small
    \begin{tabular}{l|c|c|c|c}
         &  \textbf{de-hsb} & \textbf{en-ka} & \textbf{en-kk} & \textbf{en-uk}\\\hline
        Precision & 87.08&44.8&49.3&67.4\\
        Recall &76.15&44.4&42.4&74.2\\
        F1 &81.25&44.6&45.6&70.6\\\hline
        Threshold &1.034&1.023&1.022& 1.026\\
    \end{tabular}
    \caption{The evaluation metrics on the PSM task and the respective mining thresholds.}
    \label{tab:psm}
\end{table}



 
\subsubsection{Pre-Training}
We pre-train one multilingual language model for en+ka+kk+uk and one bilingual language model for de+hsb. In one training step, the model sees a minibatch of text streams in all languages. The weights from the pre-trained language models are copied into both the encoder and the decoder of the respective bilingual NMT models. The initialized NMT model for each language pair is then further pre-trained with the denoising auto-encoding loss on the two languages until convergence. The details of the denoising task are identical to \citet{lample2018only}. 

 \subsubsection{Fine-Tuning}
We experiment with different fine-tuning strategies for unsupervised machine translation. For each language pair, all translation models are initialized with the same weights obtained in the pre-training stage described in the previous paragraph. 

\textit{IBT (baseline)} models are fine-tuned solely with the iterative back-translation loss. 

\textit{PseudoPar} models are fine-tuned with the standard supervised MT loss on our pseudo-parallel corpora.

\textit{IBT+PseudoPar} models are fine-tuned simultaneously with the iterative back-translation loss on the monolingual sentences and with the standard MT loss on the pseudo-parallel sentence pairs.

\textit{IBT+PseudoPar\/$\mapsto$IBT} models are a continuation from different checkpoints of the \textit{IBT+PseudoPar} models where the supervised MT objective is dropped and the training continues with iterative back-translation only. We experiment with different checkpoints to find the optimal point to switch the training.

\subsubsection{Evaluation}
The baseline for our approach is an improved model of \citet{conneau2019pretraining} with an extra pre-training step on the denoising task for better performance. We initialize the baseline model with the weights of a cross-lingual language model, further pre-train as a denoising autoencoder and fine-tune with iterative back-translation.

We benchmark our results against MT systems of \citet{de-gibert-bonet-etal-2022-unsupervised_mt4all}  trained as a baseline for the MT4All shared task according to the methodology of \citet{Artetxe2019effective}, and against \citet{shapiro-etal-2022-aic} who won the WMT22 de-hsb unsupervised task with a multilingual system that was pre-trained according to the mBART \citep{liu-etal-2020-multilingual-denoising} methodology and fine-tuned on synthetic texts generated by a phrase-based system.

To challenge the relevance of unsupervised MT in the world of large language models, we also translate our test sets by the GPT-3.5 Turbo model\footnote{\url{https://platform.openai.com/docs/models/gpt-3-5}} using the ChatGPT API and compare to our results. 

We measure translation quality by BLEU score using sacreBLEU\footnote{\texttt{sacrebleu -tok '13a' -s 'exp'}} \citep{sacrebleu}.

\section{Results \& Discussion}
\label{sec:disc}

\subsection{Results}

\begin{table}[]
    \centering
      \small
    \begin{tabular}{p{2.9cm}|c|c|c|c|c|c|c|c}
        & \textbf{de-hsb} & \textbf{hsb-de} & \textbf{en-ka} & \textbf{ka-en} & \textbf{en-kk} & \textbf{kk-en} & \textbf{en-uk} & \textbf{uk-en}\\\hline
        WMT22 best & 17.9 & 18.0 &-&-&-&-&-&-\\
        \hline
        ChatGPT & 6.4 & -&3.9&-&5.2&-&\textbf{25.8}&- \\
        \hline
        IBT (baseline) & 29.5 & 35.6 & 3.6 & 5.2 & 0.8 & 1.0 & 8.4 & 12.9\\
                \hline
         
        PseudoPar & 11.3 & 12.0 & 1.9 & 4.8 & 1.0 & 3.1 & 4.6 & 8.6 \\

        \hline
        IBT+PseudoPar  & 32.18 & 36.13 & 6.8 & 12.7 & 5.9 & 11.3 & 12.2 & 20.8\\
        \hspace{4em} $\mapsto$IBT & \textbf{34.94} & \textbf{39.63} & \textbf{7.7} & \textbf{14.0} & \textbf{7.2} & \textbf{12.1} & \textbf{15.7} & \textbf{23.7} \\


    \end{tabular}
    \vspace{0.5cm}

    \begin{tabular}{p{2.9cm}|c|c|c|c|c|c|c|c}
        & \textbf{de-hsb} & \textbf{hsb-de} & \textbf{en-ka} & \textbf{ka-en} & \textbf{en-kk} & \textbf{kk-en} & \textbf{en-uk} & \textbf{uk-en}\\\hline
        de Gibert Bonet (2022)  &-&-&12.0&-&6.4&-&20.8&- \\
        \hline
        IBT (baseline)&-&-  &9.0& 12.7&0.3&0.3&14.9 &12.6\\
                \hline
         
        PseudoPar&-&- & 2.1 & 6.8 & 8.0 & 11.6 & 14.6 & 13.1 \\

        \hline
        IBT+PseudoPar&-&- &11.5&22.0&\textbf{16.3}&\textbf{18.6}&\textbf{29.3}&21.7\\
       
        \hspace{4em} $\mapsto$IBT&-&-  & \textbf{15.0} & \textbf{23.5} & 9.3 & 12.7 & 27.5 & \textbf{21.8} \\


    \end{tabular}
    \caption{MT performance of our systems measured by BLEU scores on the general test set (top) and the legal test set (bottom). Compared to the WMT22 winner \citep{shapiro-etal-2022-aic}, ChatGPT, and the system trained by \citet{de-gibert-bonet-etal-2022-unsupervised_mt4all}. }
    \label{tab:test}
\end{table}

\begin{figure}
    \centering
    \includegraphics[width=13cm]{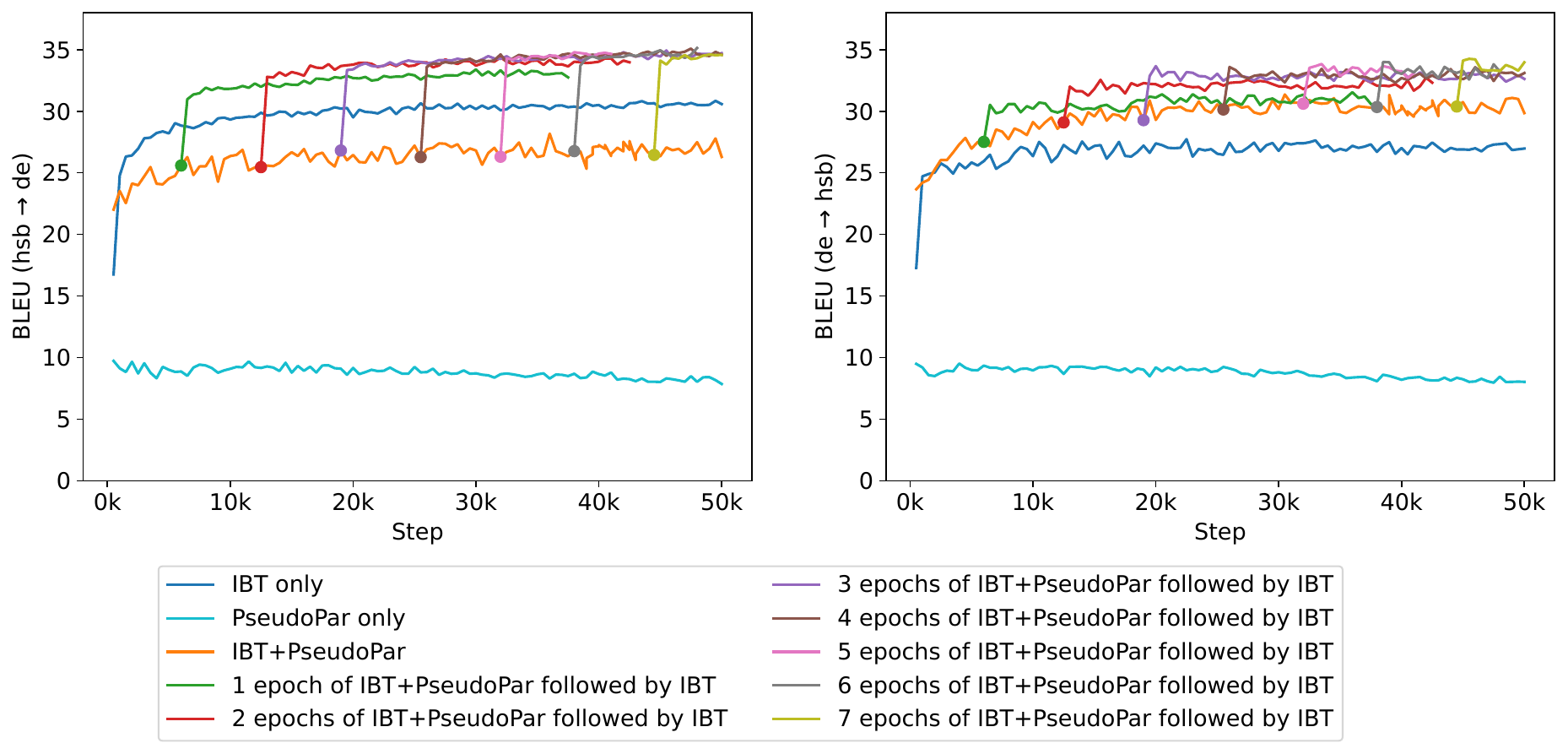}
    \caption{The development of validation BLEU scores during training. Any parallel resources were prohibited. }
    \label{fig:bleu}
\end{figure}

We observed a significant improvement in translation quality over the baseline for all translation pairs. \cref{tab:test} shows that the baseline \textit{IBT} system falls short of our proposed method by between 4.7 BLEU points (en$\xrightarrow{}$kk) and 10.7 BLEU points (uk$\xrightarrow{}$en) on the general test set. The differences on the legal test set are even more pronounced: we observe an increase of up to 14.5 BLEU over the baseline (en$\xrightarrow{}$uk). Our de$\xrightarrow{}$hsb system outperforms the WMT22 winner by 17 BLEU points. When translating from English to Kazakh, our approach reaches a BLEU score of 16.3 while the baseline which solely relies on iterative back-translation does not receive enough cross-lingual signal to start learning at all. The hybrid system by \citet{de-gibert-bonet-etal-2022-unsupervised_mt4all} which uses additional translation information from an unsupervised phrase-based system falls behind with a BLEU score of 6.4. 

The results of translation by ChatGPT from English or German into truly low-resource languages (hsb, ka, kk) are significantly worse than our results. However, after manually evaluating several translations with a zero BLEU score, we believe that the automatic metric puts ChatGPT's less literal translations at a disadvantage. ChatGPT definitely favors fluency over accuracy, but it gets zero BLEU credit even in situations when it conveys the same information in different words. Nonetheless, the en$\xrightarrow[]{}$uk translation by ChatGPT is better than all unsupervised MT systems. It must be noted that the systems cannot be directly compared to ChatGPT since its training corpus is larger and might include parallel texts. 

\subsection{Training Schedules}
\cref{fig:bleu} shows training curves with validation BLEU scores of all our de$\xleftrightarrow{}$hsb systems. We see that the \textit{IBT+PseudoPar} system trained simultaneously on back-translated and pseudo-parallel data without any special schedule outperforms the baseline for de$\xrightarrow{}$hsb but not in the opposite direction. For hsb$\xrightarrow{}$de, the baseline performance is surpassed as soon as we remove the pseudo-parallel corpus from the training.  


 We trained several de-hsb models starting from \textit{IBT+PseudoPar} after each completed epoch of 770k pseudo-parallel sentences. Upon examination of the training curves in \cref{fig:bleu}, we see an immediate increase in validation BLEU score of $\sim$0.9--4.9 BLEU points which occurred within the first 500 training steps after removing the pseudo-parallel corpus from the training. This observation confirms our hypothesis that pseudo-parallel sentence pairs aid the training in the beginning but the quality of the corpus itself poses an upper bound on the performance of the system. However, removing the corpus too early (after one or two epochs) leads to a lower final BLEU score. Therefore, we recommend to keep training the \textit{IBT+PseudoPar} model until convergence and only then switch to iterative back-translation alone \textit{IBT+PseudoPar\/$\mapsto$IBT}.

The flat \textit{PseudoPar} training curves indicate that the quality of the pseudo-parallel corpus alone is inadequate for training a functional MT system without back-translation. 

\subsection{Domain-specific MT}

Interestingly, removing the pseudo-parallel corpus from the training harms the translation quality measured on the legal test sets where the best performance for en$\xrightarrow{}$kk, kk$\xrightarrow{}$en and en$\xrightarrow{}$uk is achieved by \textit{IBT+PseudoPar}. We suspect that this is the result of the repeating terminology in the domain-specific test sets which is better handled by the \textit{IBT+PseudoPar} for some language pairs. This is consistent with the fact that the \textit{PseudoPar} system trained exclusively on pseudo-parallel data performs quite well on the en-kk and en-uk legal test set (8.0 on en$\xrightarrow[]{}$kk, 11.6 on kk$\xrightarrow[]{}$en and 14.6 on en$\xrightarrow[]{}$uk) while having poor results on the general test set (1.0 on en$\xrightarrow[]{}$kk, 3.1 on kk$\xrightarrow[]{}$en and 4.6 on en$\xrightarrow[]{}$uk). Based on our findings, we believe that utilizing pseudo-parallel sentences extracted from domain-specific monolingual corpora has the potential to enhance the training of domain-specific MT in general. However, further experiments are out of the scope of this paper.




\begin{table*}[]
    \centering
    \small
\hskip-0.2cm    \begin{tabular}{l|p{5.5cm}|p{5.5cm}| l}
\#& Upper Sorbian & German & Score \\\hline
1 & Thomas de Maizière  &  Thomas de Maizière & 1.286\\
2&Es ist ein harter Kampf, die Konkurrenz ist groß.    &   To bě napjata hra, a konkurenca bě wulka.& 1.185\\
3&Der Roman hat \textit{1200} Seiten.    &  Kniha ma \textit{300} stronow. & 1.178\\
4&Er passt zu diesem Team wie der Deckel auf den Topf. &   Wón so k mustwu hodźi kaž wěko na hornc. & 1.161 \\
5&Die größte misst über \textit{fünf Meter, die kleinste wenige Milli}meter.   &   Najkrótša měri \textit{10 cm}, najdlěša \textit{1 meter}. & 1.101\\
6& Wer Wohlstand will, braucht Wissenschaft.    &   Štóž chce \textit{něšto změnić}, \textit{trjeba sylnu wolu}.&  1.063 \\


 7 & \textit{Auch für Apple ist das iPhone wichtig.} &   \textit{Tež aleje su jara wažne.} &1.037 \\

    \end{tabular}
        \selectlanguage{english}
        
    \caption{A sample from the de-hsb mined parallel corpus. Non-matching words in italics.}
    \label{tab:excerpt}
\end{table*}

\subsection{Data quality}

The sentence pairs in the pseudo-parallel corpus are far from equivalent in meaning. As illustrated in \cref{tab:excerpt}, many of the sentences are paired because they share a named entity, a numeral (not necessarily identical), a punctuation mark, or one distinctive word. Others have a similar sentence structure, they contain a similar segment or they contain words that are somehow related, e.g. Apple/alleys (\textit{``aleje"}), although the word Apple is not the fruit in this context. On the other hand, synthetic sentences in the first training iterations are also extremely noisy, and even later they contain artifacts such as non-translated words or mistranslated named entities.

\cref{tab:onesent} shows what the back-translated and pseudo-parallel data can look like. We observed how the back-translated version of one sentence changes as the training progresses and witnessed several types of error, e.g. the German word \textit{``laufend"} is not translated at all in the initial iterations; the word ``April" remains mistranslated as ``March" (``\textit{měrc}") throughout the entire training. On the other hand, the pseudo-parallel sentence matched based on its distance from the source sentence has a similar meaning but is factually inaccurate.

We see that many of the pseudo-parallel translations are far from equivalent but it is difficult to measure the quality of the entire corpus. We measure it indirectly by the increase in BLEU score associated with introducing the corpus into the unsupervised MT training or by measuring the quality of the sentence encoder used for creating the corpus. To be able to evaluate the precision/recall of the sentence encoder, we have to control the number of parallel sentences hidden in the input corpora. However, in real-life scenarios, the level of comparability of two monolingual corpora is never known precisely. If the monolingual corpora provided for unsupervised translation come from a different domain and contain dissimilar sentences, the model has no good candidates to find. This poses a challenge especially when setting the correct mining threshold for the monolingual corpora at hand. 

 It is not clear what are the attributes of the pseudo-parallel corpus that the unsupervised MT training benefits from the most. We believe that the benefits of training on such noisy data are twofold: 1) the perfect matches are a valuable source of correct supervision, and 2) the abundant less-than-perfect matches still introduce a new translation signal which can help the model leave a suboptimal situation which we often observe during back-translation when the model learns to mistranslate a word and never forgets it.



\begin{table}[]
    \centering
    \small
     
\hskip-0.2cm    \begin{tabular}{l | p{10cm}l}

SRC & Ich musste mich laufend weiterbilden, und so legte ich im April 1952 die erste und ein Jahr darauf die zweite Lehramtsprüfung ab. \\
REF & Dyrbjach so běžnje dale kwalifikować, a tak złožich w aprylu 1952 prěnje a lěto po tym druhe wučerske pruwowanje. \\
PseudoPar & \textit{Hańža Winarjec-Orsesowa} wotpołoži prěnje wučerske pruwowanje \textit{w lěće 1949 a druhe w lěće} 1952.\\
IBT @ 500 & Dyrbjach so \textit{laufend} dale \textit{kubłać}, a tak \textit{legte w měrcu} 1952 \textit{prěnje a lěto na to druhe Lejnjanske pruwowanje ab.} \\
IBT @ 3000 & Dyrbjach so běžnje dale \textit{kubłać}, a tak w \textit{měrcu} 1952 prěnju a lěto na to druhu \textit{lektoratu serbšćiny wotpołožichmy.} \\
IBT @ 10000 & Dyrbjach so běžnje dale \textit{kubłać}, a tak wotpołožich w \textit{měrcu} 1952 prěnju a lěto \textit{na} to druhu \textit{lektoratu.} \\
     
    \end{tabular}
        \selectlanguage{english}
    \caption{A sample sentence translated by the IBT model after 500, 3,000 and 10,000 training steps compared to the closest neighbor of such sentence from the bilingual sentence space (PseudoPar). The mistranslated words are indicated in italics.}
    \label{tab:onesent}

\end{table}

\section{Conclusion}
\label{sec:concl}

We have demonstrated the benefits of MT training on pseudo-parallel data in situations when true parallel data is not available. While the pseudo-parallel corpus alone does not reach sufficient quality for standard supervised MT training, it works well in combination with iterative back-translation. It is optimal to train the model until convergence on both pseudo-parallel and synthetic sentence pairs, remove the pseudo-parallel corpus and continue training with iterative back-translation only.

Incorporating similar sentence pairs into the standard unsupervised MT training  increases translation quality across all evaluated language pairs with an improvement of up to 14.5 BLEU over the baseline trained without pseudo-parallel data and 8.5 BLEU over a hybrid unsupervised system (en$\xrightarrow[]{}$uk). 
Furthermore, we observed that in some situations (en$\xleftrightarrow{}$kk), the iterative back-translation becomes trapped in a suboptimal state where no learning occurs. Introducing pseudo-parallel data can rescue the model from this state and trigger the learning process.

After evaluating our approach on a legal test set, we believe that training on pseudo-parallel sentences could be particularly useful for domain-specific unsupervised MT. If we have two in-domain monolingual corpora at hand, parallel corpus mining is an efficient strategy to retrieve translation information.

The pseudo-parallel corpus helps the training despite being noisy. We hypothesize that while exact translations help the model find correct correspondences, also the noise can introduce new information and prevent the model from memorizing some of the artifacts of back-translated sentences.  We leave it up to future research to evaluate whether a cleaner but smaller corpus would bring even larger gains.

\section*{Acknowledgements}

This research was partially supported by the grant 19-26934X  (NEUREM3)  of  the  Czech  Science Foundation and by the SVV project number 260~698 of the Charles University.



\small
\bibliographystyle{apalike}
\bibliography{mtsummit2023,bibliography}

\end{document}